\documentclass[letterpaper]{article} 
\usepackage{aaai2026}  
\usepackage{times}  
\usepackage{helvet}  
\usepackage{courier}  
\usepackage[hyphens]{url}  
\usepackage{graphicx} 
\usepackage{natbib}  
\usepackage{caption} 
\usepackage{algorithm}
\usepackage{algorithmic}
\usepackage{booktabs}
\usepackage{multirow}  
\usepackage{amsmath}
\usepackage{cleveref}
\usepackage{subcaption}

\usepackage{newfloat}
\usepackage{listings}
\DeclareCaptionStyle{ruled}{labelfont=normalfont,labelsep=colon,strut=off} 
\floatstyle{ruled}
\newfloat{listing}{tb}{lst}{}
\floatname{listing}{Listing}
\title{MultiEditor: Controllable Multimodal Object Editing for Driving Scenarios Using 3D Gaussian Splatting Priors}
\author{
    Shouyi Lu\textsuperscript{\rm 1} Zihan Lin\textsuperscript{\rm 2}\thanks{Project leader.} Chao Lu\textsuperscript{\rm 2} Huanran Wang\textsuperscript{\rm 2} Guirong Zhuo\textsuperscript{\rm 1}\thanks{Corresponding author.} Lianqing Zheng\textsuperscript{\rm 1}\\
}
\affiliations{
    \textsuperscript{\rm 1}School of Automotive Studies, Tongji University\\
    \textsuperscript{\rm 2}Mach Drive

}

\usepackage{bibentry}

\begin{document}
	
\maketitle

\begin{abstract}
	Autonomous driving systems rely heavily on multimodal perception data to understand complex environments. However, the long-tailed distribution of real-world data hinders generalization, especially for rare but safety-critical vehicle categories. To address this challenge, we propose MultiEditor, a dual-branch latent diffusion framework designed to edit images and LiDAR point clouds in driving scenarios jointly. At the core of our approach is introducing 3D Gaussian Splatting (3DGS) as a structural and appearance prior for target objects. Leveraging this prior, we design a multi-level appearance control mechanism—comprising pixel-level pasting, semantic-level guidance, and multi-branch refinement—to achieve high-fidelity reconstruction across modalities. We further propose a depth-guided deformable cross-modality condition module that adaptively enables mutual guidance between modalities using 3DGS-rendered depth, significantly enhancing cross-modality consistency. Extensive experiments demonstrate that MultiEditor achieves superior performance in visual and geometric fidelity, editing controllability, and cross-modality consistency. Furthermore, generating rare-category vehicle data with MultiEditor substantially enhances the detection accuracy of perception models on underrepresented classes.
\end{abstract}

\section{Introduction}
Autonomous driving systems rely on diverse sensors to perceive their surroundings. Among them, LiDAR and cameras are the primary modalities, capturing point clouds and RGB images that offer complementary geometric and semantic information crucial for scene understanding. Despite significant advances in multimodal perception \cite{yolo,pointpillars,Zhao_2024_CVPR,chae2024towards}, the performance of existing models remains heavily dependent on large-scale, balanced datasets. In practice, real-world driving data often exhibit a pronounced long-tailed distribution: common vehicle categories are vastly overrepresented, while rare yet safety-critical classes—such as road rollers and excavators—are severely underrepresented. This data imbalance hinders generalization and undermines the robustness of perception models in long-tail and edge-case scenarios.
\begin{figure}[t]
	\centering
	\includegraphics[width=0.9\linewidth]{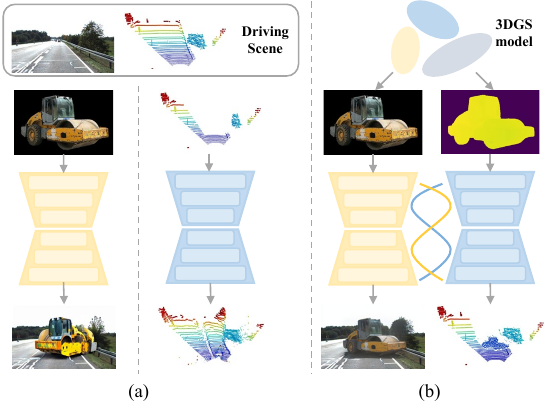}
	\caption{(a) Editing image and point cloud separately using independent single-modal models \cite{anydoor,rangeldm}. (b) Our proposed MultiEditor leverages 3DGS priors to jointly edit image and LiDAR data, enhancing geometric and appearance accuracy and ensuring cross-modality consistency.}
	\label{fig1}
\end{figure}

To mitigate the challenge of limited data diversity, recent research \cite{mobi, subjectdrive, driveeditor, rangeldm, streetcrafter, omni} has explored the use of Latent Diffusion Models (LDMs) \cite{ldm} to synthesize tailored driving scenarios. However, most existing approaches are restricted to single-modality editing—focusing either on images or point clouds—and lack a unified framework for joint multimodal editing. Naively combining separate single-modal pipelines often leads to substantial cross-modality inconsistencies in geometry and appearance (see Figure~\ref{fig1}(a)). Furthermore, image editing methods typically rely on dataset-derived object priors, offering limited flexibility in manipulating object pose and viewpoint. Likewise, point cloud editing methods frequently depend on scene-level masks and overlook object-level structural priors, reducing editing precision and controllability.

In this paper, we introduce MultiEditor, a novel framework for joint editing of LiDAR and image data, as shown in Figure~\ref{fig1}(b). MultiEditor is the first framework to incorporate 3D Gaussian Splatting (3DGS) \cite{3dgs} as a unified prior for both the appearance and structure of target objects. By leveraging 3DGS’s capability to render RGB images and depth maps from arbitrary viewpoints, our framework enables precise and controllable manipulation of object position and orientation. Architecturally, MultiEditor employs a dual-branch latent diffusion framework, comprising two dedicated generative pathways for point clouds and images, and unifies them via a shared 3DGS-based object representation.

To enable high-fidelity object editing, MultiEditor integrates 3DGS-rendered information into each branch and introduces a multi-level appearance control mechanism comprising pixel-level preservation through image pasting, semantic-level guidance via embedding features, and a multi-branch joint optimization strategy. To ensure cross-modality consistency, we propose a depth-guided deformable condition module that leverages 3DGS-rendered depth priors to facilitate latent-space feature alignment and mutual guidance across modalities. Additionally, to address challenges such as depth noise, we introduce a deformable attention-based \cite{dfa} multimodal interaction mechanism that adaptively establishes cross-modality correspondences without requiring precise geometric registration, thereby mitigating performance degradation caused by misalignment.

We validate MultiEditor's effectiveness through comprehensive evaluations on public datasets. Experimental results show our method surpasses state-of-the-art single-modality approaches in visual and geometric fidelity, editing controllability, and cross-modality consistency. MultiEditor is the first framework capable of flexibly editing atypical vehicles, offering a novel solution for enhancing perception robustness in long-tail scenarios.  

Our contributions are summarized as follows.

\begin{itemize}
	\item We propose MultiEditor, the first dual-branch multimodal latent diffusion framework that integrates 3DGS as a unified prior for object appearance and structure, enabling flexible, precise, and consistent joint editing of point clouds and images.
	\item We introduce a depth-guided deformable cross-modality condition module that leverages 3DGS-rendered depth and spatial transformations to align latent-space features and enable mutual guidance across modalities.
	\item Extensive experiments demonstrate that MultiEditor achieves superior performance in visual and geometric fidelity, editing controllability, cross-modality consistency, and effectiveness in downstream perception tasks.
\end{itemize}

\section{Related Work}
\subsection{Generation of Driving Scenarios}
The growing demand for diverse multimodal data in autonomous driving has accelerated progress in synthetic data techniques. Most existing approaches focus on single-modality generation by leveraging generative models. In image synthesis, prior work \cite{magicdrive,drivedreamer,drivedreamer2,panacea,uniscene,dualdiff} has achieved controllable generation by conditioning pre-trained diffusion models on structured inputs such as road layouts and bounding boxes. For 3D point clouds, recent advances \cite{rangeldm,lidardm,r2dm,text2lidar} project point clouds into range images, enabling efficient synthesis via 2D diffusion architectures. Despite notable success in modeling marginal distributions of single-modality data, existing methods largely overlook the mutual dependencies between modalities necessary to capture the complexity of driving scenes. To fill this gap, X-Drive \cite{xdrive} proposes interaction mechanisms between heterogeneous diffusion models, facilitating joint 2D–3D generation via cross-modality interaction. \textit{Despite these advances in full-scene synthesis, existing approaches remain constrained by global generation paradigms and struggle to support fine-grained, object-level editing.}
\subsection{Editing of Driving Scenarios}
Unlike scene generation, driving scene editing emphasizes fine-grained manipulation of specific objects within scenarios. R3D2 \cite{r3d2} performs asset insertion by leveraging synthetic assets and 3DGS-reconstructed scenes, employing a single-step diffusion model for scene-consistent inpainting. PbE \cite{pbe} proposes an exemplar-driven image editing method using diffusion models, enabling fine-grained semantic manipulation through self-supervised training and boundary artifact suppression. AnyDoor \cite{anydoor} introduces a diffusion-based framework augmented with identity and detail feature extractors, supporting flexible object synthesis at user-defined locations. SubjectDrive \cite{subjectdrive} design a video generation architecture that integrates a subject prompt adapter, a visual adapter, and augmented temporal attention, substantially improving controllability over generated subjects. Liang et al. \cite{driveeditor} combine 3D bounding box priors with novel view synthesis to achieve object-level editing in autonomous driving videos. RangeLDM \cite{rangeldm} proposes a diffusion-based approach for synthesizing range-view LiDAR point clouds, which also supports mask-conditioned point cloud editing. MObI \cite{mobi} develops a unified diffusion model for camera–LiDAR modalities to enable joint cross-sensor editing. GenMM \cite{genmm} incorporates depth estimation into video diffusion models to support cross-modality editing propagation. \textit{Despite these advancements, existing methods remain constrained by predefined vehicle categories in the training data, provide limited control over object poses, and lack explicit mechanisms for cross-modality interaction. To overcome these limitations, we introduce a multimodal editing framework built upon 3DGS, leveraging a 3DGS-based vehicle template library to enable physically plausible insertions of diverse vehicle types with complex poses. In addition, we design a depth-guided cross-modality condition module that utilizes 3DGS-rendered depth to facilitate end-to-end joint optimization of images and point clouds.}   
\section{The Proposed Method}
\begin{figure*}[t]
	\centering
	\includegraphics[width=0.9\linewidth]{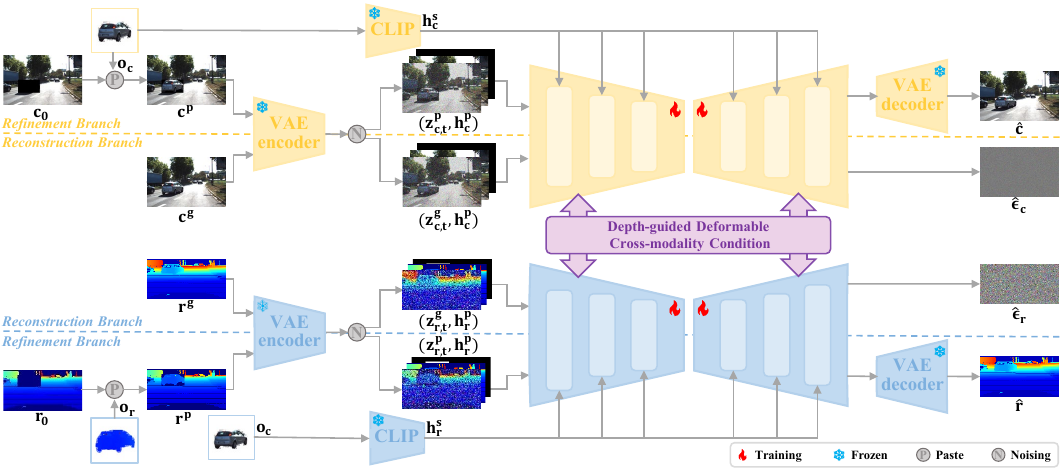}
	\caption{Overview of the proposed MultiEditor framework. A dual-branch diffusion model is employed to edit multimodal data. Each branch incorporates a multi-level appearance control mechanism for fidelity, while a cross-modality condition module enhances consistency between modalities.}
	\label{fig2}
\end{figure*}
\subsection{Multimodal Joint Editing via Diffusion Models}  
\label{3.1}
Multimodal joint editing aims to precisely manipulate target objects across images and point clouds by jointly modeling and integrating cross-modality information. For each modality, we employ a conditional diffusion process to insert the target object at the location specified by the scene mask, ensuring seamless integration with the surrounding context. Moreover, both spatial and semantic consistency must be maintained across modalities.

Formally, let $c$ denote an image, $r$ the range-view representation of the point cloud, and $m_c$ and $m_r$ the corresponding scene masks. We begin by masking the regions of interest in both modalities, defined as:
\begin{equation}
	\label{eqs_1}
	c_0 = m_c \odot c, \quad \quad r_0 = m_r \odot r,
\end{equation}
where $\odot$ denotes element-wise multiplication. A shared noise schedule $\{\beta_t\}_{t=1}^T$ is applied to both masked inputs, and independent forward diffusion processes are constructed for each modality:
\begin{equation}
	\label{eqs_2}
	\begin{aligned}
		q(r_t \mid r_{t-1}) &= \mathcal{N}(r_t; \sqrt{1 - \beta_t} r_{t-1}, \beta_t \mathbf{I}), \\
		q(c_t \mid c_{t-1}) &= \mathcal{N}(c_t; \sqrt{1 - \beta_t} c_{t-1}, \beta_t \mathbf{I}).
	\end{aligned}
\end{equation}

In the reverse process, we introduce two denoising models, $\epsilon_{\theta_r}$ and $\epsilon_{\theta_r}$, to estimate the noise $\mathbf{\epsilon}_r$ and $\mathbf{\epsilon}_c$ added to the range image and RGB image separately. Under the guidance of target objects $o_r$ and $o_c$, each model progressively integrates the objects into the background. To enhance cross-modality consistency, the noise prediction for each modality incorporates complementary cues from the other modality. Intuitively, intermediate features from the diffusion process can serve as conditional inputs to guide the denoising of the other modality. The denoising models can thus be formulated as:
\begin{equation}
	\label{eqs_3}
	\begin{aligned}
		\hat{\epsilon}_{r} &= \epsilon_{\theta_r}(r_t, c_{CR}(c_t, t), o_r, t),\\
		\hat{\epsilon}_{c} &= \epsilon_{\theta_c}(c_t, c_{RC}(r_t, t), o_c, t).
	\end{aligned}
\end{equation}

A well-defined spatial correspondence between images and point clouds is established using the sensor extrinsics $T(\cdot)$ and the target object's depth $D$. However, previous methods \cite{mobi} lack the depth of inserted objects. To address this limitation, we incorporate the 3DGS representation of the target object to render depth maps and facilitate cross-modality interaction. Accordingly, the cross-modality conditional encoders $c_{CR}(\cdot)$ and $c_{RC}(\cdot)$ are defined as:
\begin{equation}
	\label{eqs_4}
	\begin{aligned}
		c_{CR}(c_t, t) &= T_{CR}(\epsilon'_{\theta_c}(c_t, t), D), \\
		c_{RC}(r_t, t) &= T_{RC}(\epsilon'_{\theta_r}(r_t, t), D), 
	\end{aligned}
\end{equation}
where $\epsilon'_{\theta_c}(\cdot)$ and $\epsilon'_{\theta_r}(\cdot)$ represent intermediate features extracted from the image and point cloud denoising models, respectively. 
In this case, we rewrite the denoising models in Eq.~(\ref{eqs_3}) as follows:
\begin{equation}
	\label{eqs_5}
	\begin{aligned}
		\hat{\epsilon}_{r} &= \epsilon_{\theta_r}\left(r_t, T_{CR}(\epsilon'_{\theta_c}(c_t, t), D), o_r, t\right), \\
		\hat{\epsilon}_{c} &= \epsilon_{\theta_c}\left(c_t, T_{RC}(\epsilon'_{\theta_r}(r_t, t), D), o_c, t\right). 
	\end{aligned}
\end{equation}

They can be trained with a joint multi-modality objective function $\mathcal{L}_{DM-M}$.
\begin{equation}
	\label{eqs_6}
	\mathcal{L}_{DM-M} = \mathcal{L}_{DM-C} + \mathcal{L}_{DM-R},
\end{equation}
where $\mathcal{L}_{DM-C}$ and $\mathcal{L}_{DM-R}$ represent the loss terms for the image and point cloud modalities, respectively.

\subsection{Dual-Branch Joint Editing Framework} 
\label{3.2}
As illustrated in Figure~\ref{fig2}, our framework is formulated based on Eq.~(\ref{eqs_5}) and comprises two denoising models: $\epsilon_{\theta_c}(\cdot)$ for image editing and $\epsilon_{\theta_r}(\cdot)$ for range image editing.
\subsubsection{Latent Diffusion Model for Image Editing} \label{idm}
In the image branch, we employ an LDM tailored for image editing tasks. In addition to cross-modality guidance from the range image branch, we integrate three condition mechanisms to enable high-fidelity synthesis of the target object: (i) pixel-level preservation through image pasting, (ii) semantic-level guidance via embedding features, and (iii) a multi-branch joint optimization strategy.

\textbf{Pixel-level detail preservation.}
To preserve the fine-grained appearance detail of the target object, we adopt a simple yet effective paste-based strategy. During training, a pretrained image segmentation model \cite{sam,groundingdino} is used to extract the target object from the scene and paste it into the Region of Interest (ROI) specified by the scene mask. During inference, the 3DGS-rendered image of the target object is pasted into the ROI. The pasted image $c^p$, after being compressed by a variational autoencoder (VAE) \cite{vae}, is concatenated with the downsampled scene mask $m_c$ to generate the pixel-level conditional features $h_c^p$.
\begin{equation}
	\label{eqs_7}
	h_c^p = \text{Concat}(\text{VAE}(c^p), m_c).
\end{equation}

This feature is then fused into the diffusion process through channel-wise concatenation with the latent representation.

\textbf{Semantic consistency maintenance.}
To enhance semantic preservation and contextual alignment, we use a frozen image encoder coupled with a trainable $\text{MLP}_c$ to extract global semantic features from the target object. Following pioneering work \cite{pbe}, we adopt CLIP \cite{clip} as the image encoder due to its strong capability in capturing high-level semantic representations. The resulting CLIP embedding is a conditional signal integrated into the image denoising model via a cross-attention module. The semantic condition $h_c^s$ is formulated as:
\begin{equation}
	\label{eqs_8}
	h_c^s = \text{MLP}_c(\text{CLIP}(o_c)).
\end{equation}

\textbf{Multi-branch optimization.}
To more effectively utilize the coarse editing results of the pasted image, inspired by DCI-VTON \cite{gou2023taming}, we adopt a dual-branch optimization strategy, comprising a reconstruction branch and a refinement branch, to train the denoising network $\epsilon_{\theta_c}(\cdot)$. The reconstruction branch synthesizes the driving scene conditioned on the reference input, promoting semantic alignment and structural consistency. In parallel, the refinement branch refines the pasted image by enhancing fine-grained details, thereby enabling high-fidelity data synthesis. Specifically, the reconstruction branch synthesizes images by applying a forward diffusion followed by a reverse denoising process on the real image $c^g$. In line with standard LDM designs, $c^g$ is first encoded into a latent representation $z_{c,0}^g$ via a VAE. Gaussian noise is then added according to Eq.~(\ref{eqs_2}), and the resulting noisy latent is concatenated with the pixel-level condition $h_c^p$ as input to the denoising network. In parallel, the refinement branch targets the masked region of the pasted image $c^p$. The latent representation $z_{c,0}^p$, obtained from VAE encoding of $c^p$, is likewise perturbed with Gaussian noise and concatenated with the pixel-level condition $h_c^p$.

The reconstruction branch guides the denoising model by supervising its noise prediction, with the objective function $\mathcal{L}_{recon-C}$ defined as:
\begin{equation}
	\label{eqs_9}
	\mathcal{L}_{recon-C} = \left\| \epsilon_c - \epsilon_{\theta_c}(z_{c,t}^g, h_c^p, h_c^s, h_c^r, t) \right\|_2^2,
\end{equation}
where $h_c^r$ denotes the cross-modality condition. The refinement branch obtains the denoised latent representation $\hat{z}_{c,0}^g$ by reversing the noising process using the predicted noise. This latent is then decoded into an image $\hat{c}=D_{VAE}(\hat{z}_{c,0}^g)$. To supervise the refinement, we compute both the L2 loss and the perceptual loss \cite{perceptual_loss} between the reconstructed image $\hat{c}$ and the real image $c^g$, defined as:
\begin{equation}
	\label{eqs_10}
	\mathcal{L}_{refine-C} = \|c^g - \hat{c}\|_2^2 + \sum_{m=1}^{5} \|\phi_m(\hat{c}) - \phi_m(c^g)\|_1.
\end{equation}
where $\phi_m(\cdot)$ indicates the $m$-th feature map in a VGG-19 \cite{vgg16} network pre-trained on ImageNet \cite{imagenet}.
Accordingly, the overall objective function for the image branch is formulated as:
\begin{equation}
	\label{eqs_11}
	\mathcal{L}_{DM-C} = \mathcal{L}_{recon-C} + \lambda_{refine-C} \mathcal{L}_{refine-C},
\end{equation}
where $\lambda_{refine-C}$ serves as a hyperparameter to balance the contributions of the two loss components.
\subsubsection{Latent Diffusion Model for Range Image Editing} \label{rdm}
Given the structural similarity between range images and RGB images, we adapt the LDM from the image branch to accommodate the characteristics of range data. While retaining the semantic condition signal and overall optimization strategy, we customize the VAE and the pixel-level condition module to better model the spatial and geometric information in point cloud data.

We adopt the standard VAE training paradigm, jointly optimizing the encoder and decoder by maximizing the ELBO \cite{stochastic} to learn a compact latent representation of range images. To mitigate the blurriness typically introduced by reconstruction loss, we further incorporate an adversarial discriminator \cite{isola2017image,ldm}, encouraging sharper outputs and improved structural fidelity.

\textbf{Pixel-level spatial preservation in the range view.} To preserve the spatial structure of the target object, we also adopt a paste-based strategy in the point cloud branch. Due to the difficulty of acquiring LiDAR point clouds for the target object under diverse poses and positions, we leverage the depth rendering capabilities of 3DGS to synthesize dense depth maps that serve as spatial priors, effectively capturing the geometric structure of the object.
During training, we interpolate sparse LiDAR data to generate a dense depth map and extract the target object’s depth using segmentation masks produced by the image branch. The extracted depth is then pasted into the ROI specified by the scene mask. To reduce boundary artifacts caused by interpolation noise, we apply median filtering to suppress unreliable depth estimates. During inference, we directly use depth maps rendered by the 3DGS model. Analogous to the image branch, we construct the pixel-level condition feature $h_r^p$ by concatenating the VAE-compressed pasted range image $r^p$ with the downsampled scene mask $m_r$:
\begin{equation}
	\label{eqs_12}
	h_r^p = \text{Concat}(\text{VAE}(r^p), m_r).
\end{equation}

This feature is integrated into the diffusion model of the range image branch via channel-wise concatenation.
\begin{figure}[t]
	\centering
	\includegraphics[width=0.9\linewidth]{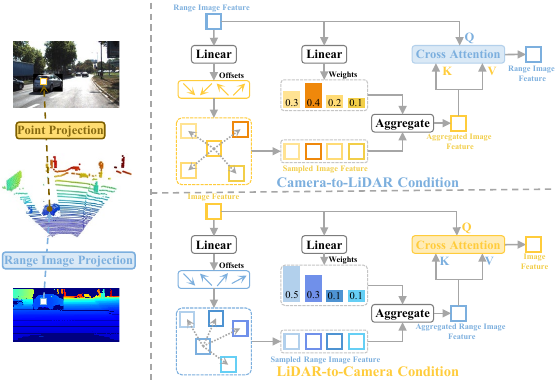}
	\caption{Cross-modality condition module. We perform bidirectional conditioning between LiDAR and camera modalities on latent representations, guided by depth priors and spatial transformations.}
	\label{fig3}
\end{figure}
\subsection{Cross-Modality Condition Module} \label{sec:cmc}
The key to multimodal editing is boosting the cross-modality consistency, which potentially relies on cross-modality conditions. While the depth prior of the target object and the camera projection inherently establish a spatial correspondence between image and point cloud modalities, this alignment is often inaccurate due to depth estimation errors. To address these challenges, we introduce a depth-guided deformable condition module, illustrated in Figure~\ref{fig3}. This module initially establishes a coarse modality alignment by leveraging the depth prior and geometric transformations, followed by a deformable cross-attention mechanism that adaptively retrieves local features from the complementary modality, thereby generating cross-modality control signals.
\subsubsection{Camera-to-LiDAR condition} 
Given a location $(\phi, \theta)$ in the range image latent $z_r$ with corresponding range value $r$, we first convert it into 3D Cartesian coordinates:
\begin{equation}
	\label{eqs_13}
	\begin{aligned}
		x &= r \cdot \cos(\phi) \cdot \sin(\theta), \\
		y &= r \cdot \cos(\phi) \cdot \cos(\theta), \\
		z &= r \cdot \sin(\phi).
	\end{aligned}
\end{equation}

Subsequently, we project the 3D point onto the image plane using the camera intrinsic matrix $K$ and the LiDAR-to-camera extrinsic transformation $T_{CR} = [R_{CR} \mid t_{CR}]$:
\begin{equation}
	\label{eqs_14}
	\begin{bmatrix}
		u \quad v \quad 1
	\end{bmatrix}^\top = \frac{K}{d} \left( R_{CR} \begin{bmatrix} x \quad y \quad z \end{bmatrix}^\top + t_{CR} \right),
\end{equation}
where $(u,v)$ denotes the pixel location in the image and $d$ is a normalization factor. Based on this geometric projection, we employ a deformable cross-attention module to adaptively sample local reference features around $(u,v)$ from the image latent $z_c$, forming a camera-to-LiDAR control signal: 
\begin{equation}
	\label{eqs_15}
	h_{r}^c = \text{CrossDAttn}(z_r(\phi, \theta), z_c, (u, v)).
\end{equation}

The output at $(\phi, \theta)$ is then computed as:
\begin{equation}
	\label{eqs_16}
	z_r^{\text{out}}(\phi, \theta) = z_r(\phi, \theta) + \tanh(\alpha_r) \cdot h_{r}^c,
\end{equation}
where $\alpha_r$ is a zero-initialization gate.
\subsubsection{LiDAR-to-camera condition} 
Based on the established correspondence between the range image and the RGB image, we inject range features into the image branch. For a pixel $(u,v)$ in the image latent $z_c$, corresponding to a range image location $(\phi, \theta)$, the cross-modality condition is formulated as:
\begin{equation}
	\label{eqs_17}
	h_{c}^r = \text{CrossDAttn}(z_c(u,v), z_r, (\phi, \theta)).
\end{equation}

We then update the image latent representation as:
\begin{equation}
	\label{eqs_18}
	z_c^{\text{out}}(u,v) = z_c(u,v) + \tanh(\alpha_c) \cdot h_{c}^r,
\end{equation}
where $\alpha_c$ is a zero-initialization gate. By exploiting the 2D–3D correspondences, our condition module adaptively enhances local cross-modality consistency.
\begin{table*}[t]
	\centering
	\small
	\resizebox{0.8\linewidth}{!}{  
		\begin{tabular}{c c c c c c c c}
			\toprule
			\multirow{2}{*}{Modality} & \multirow{2}{*}{Method} & \multicolumn{3}{c}{Image quality} & \multicolumn{2}{c}{Point clouds quality} & Cross-modality \\
			\cmidrule(lr){3-5} \cmidrule(lr){6-7} \cmidrule(lr){8-8}
			& & FID $\downarrow$ & LPIPS $\downarrow$ & CLIP-I $\uparrow$ & CD $\downarrow$ & FPD $\downarrow$ & DAS $\downarrow$\\
			\midrule
			\multirow{3}{*}{C} 
			& SD                 & 67.92  & 0.3183 & 0.7805 & --       & --       & -- \\
			& PbE                & 54.28  & 0.3293 & 0.7875 & --       & --       & -- \\
			& AnyDoor            & 26.45  & 0.2680 & 0.7925 & --       & --       & -- \\
			\midrule
			L 
			& RangeLDM           & --     & --     & --     & 33.23    & 293.45   & -- \\
			\midrule
			\multirow{2}{*}{C+L} 
			& AnyDoor+RangeLDM   & --     & --     & --     & --       & --       & 11.34 \\
			& MultiEditor(ours)      & \textbf{25.07} & \textbf{0.1477} & \textbf{0.8063} & \textbf{1.65} & \textbf{97.49} & \textbf{3.16} \\
			\bottomrule
	\end{tabular}}
	\caption{Quantitative comparison with driving data editing algorithms. For each column, the best value is highlighted by \textbf{bold}.}
	\label{tab:quantitative_results}
\end{table*}
\begin{figure*}[t]
	\centering
	\includegraphics[width=0.9\linewidth]{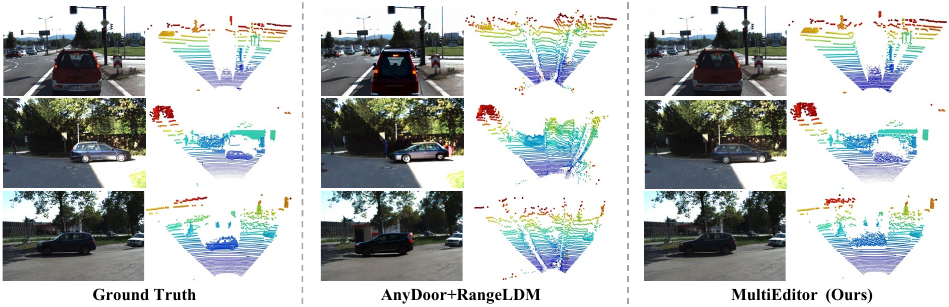}
	\caption{Editing results on regular vehicles. MultiEditor achieves better appearance and geometric fidelity than the baseline.}
	\label{fig4}
\end{figure*}
\begin{figure}[t]
	\centering
	\includegraphics[width=0.99\linewidth]{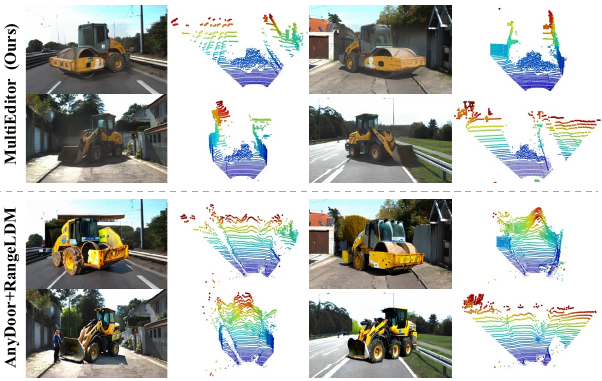}
	\caption{Editing results on atypical vehicles.}
	\label{fig5}
\end{figure}

\section{Experiments}
\subsection{Experimental Setups}
\subsubsection{Dataset and Data Construction}
We train MultiEditor with a reconstruction-based strategy. Specifically, we simulate editing scenarios by deliberately occluding the same target object in image and point cloud modalities. The model is trained to reconstruct the occluded regions conditioned on the target object accurately. As no publicly available dataset is designed explicitly for joint multimodal editing, we construct a dedicated dataset derived from the widely adopted KITTI benchmark \cite{kitti}. Details of the dataset construction process are provided in the supplementary materials.
\subsubsection{Evaluation Metrics}
For joint multimodal data editing, we evaluate both the realism of single-modality data within masked regions and the cross-modality consistency between them. For image modality, we use the Fr\'echet Inception Distance (FID) \cite{gans} for visual realism, Learned Perceptual Image Patch Similarity (LPIPS) \cite{lpips} for perceptual similarity, and CLIP-based image similarity (CLIP-I) \cite{clipscore} for semantic alignment. For the point cloud modality, we adopt Chamfer Distance (CD) \cite{cd} and Fr\'echet Point Cloud Distance (FPD) \cite{fpd} to quantify spatial and perceptual accuracy. Additionally, we follow the X-Drive \cite{xdrive} protocol and employ the Depth Alignment Score (DAS) to evaluate the consistency between generated images and point clouds.
\subsubsection{Baselines}
For single-modality editing, we compare state-of-the-art editing algorithms for images, \textit{i.e.} Stable Diffusion Inpainting (SD) \cite{ldm}, Paint-by-Example (PbE) \cite{pbe}, AnyDoor \cite{anydoor}, and for point clouds, \textit{i.e.} RangeLDM \cite{rangeldm}. Furthermore, following the strategy adopted in X-Drive \cite{xdrive}, we combine AnyDoor \cite{anydoor} and RangeLDM \cite{rangeldm} respectively for images and point clouds as a multi-modality baseline.
\subsubsection{Training Setup}
Our MultiEditor has a dual-branch architecture. During training, both branches are initialized with pretrained weights from the PbE model, while the newly introduced parameters are initialized randomly. The training process is divided into five stages, with detailed strategies in the supplementary materials.

\subsection{Quantitative Results}
Table~\ref{tab:quantitative_results} presents quantitative comparisons with several baseline methods. For image quality, MultiEditor achieves the best performance across all metrics, reducing FID to 25.07 and LPIPS to 0.1477, indicating clear gains in visual fidelity and perceptual similarity. It also obtains a CLIP-I score of 0.8063, demonstrating strong semantic alignment with the target objects. For point cloud quality, our approach significantly outperforms RangeLDM, lowering CD from 33.23 to 1.65 and FPD from 293.45 to 97.49, reflecting substantial improvements in geometric accuracy and structural consistency. Regarding cross-modality consistency, MultiEditor achieves a DAS of 3.16, notably better than the baseline combination of AnyDoor and RangeLDM. This result validates the effectiveness of our cross-modality condition module in improving cross-modality consistency for joint editing tasks.
\subsection{Qualitative Analysis}
\subsubsection{Editing of Regular Vehicles}
Figure~\ref{fig4} presents multimodal editing results on regular vehicles. For the image modality, while existing editing methods maintain semantic consistency, they often struggle to preserve fine-grained geometric structures, resulting in inaccurate shape reconstruction, incorrect pose estimation, and unnatural background blending. MultiEditor generates accurate and context-consistent image content, exhibiting superior visual fidelity and editing controllability. For the point cloud modality, mainstream approaches such as RangeLDM lack object-level priors, leading to geometric artifacts and semantic drift. Benefiting from the depth and structural priors provided by 3DGS, MultiEditor generates point clouds with clear structure.
\subsubsection{Editing of Atypical Vehicles}
Figure~\ref{fig5} presents multimodal editing results on atypical vehicles under diverse poses and environments. We construct 3DGS models of atypical vehicles using the 3DRealCar dataset \cite{3drealcar}. In the image modality, existing methods often produce distorted or semantically inconsistent editing results on irregular structures. In contrast, MultiEditor preserves complex geometry and generates visually coherent content that blends naturally with the background. For point clouds, existing methods struggle with structural fidelity due to the lack of geometric priors. With 3DGS guidance, MultiEditor synthesizes complete and semantically aligned point clouds.
\subsubsection{Cross-Modality Consistency}
\begin{figure}[t]
	\centering
	\includegraphics[width=0.95\linewidth]{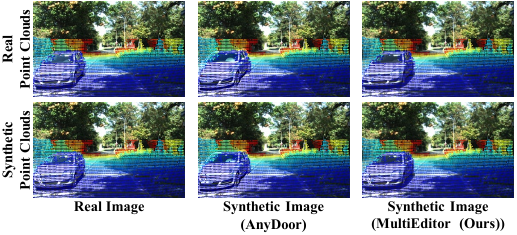}
	\caption{Qualitative results of cross-modality consistency.}
	\label{fig6}
\end{figure}
To qualitatively evaluate cross-modality consistency, we visualize the results of joint editing by projecting real and MultiEditor-generated point clouds onto three types of images: real images, images generated by AnyDoor, and those generated by MultiEditor. As illustrated in Figure~\ref{fig6}, MultiEditor yields better geometric alignment between point clouds and images than AnyDoor. Furthermore, projections of generated point clouds by MultiEditor onto their corresponding generated images exhibit noticeably superior alignment than those using real point clouds, underscoring its ability to produce structurally and visually consistent multimodal outputs.

\begin{table}[t]
	\centering
	\small
	\resizebox{0.9\linewidth}{!}{
		\begin{tabular}{cccc}
			\toprule
			Methods & LPIPS $\downarrow$ & FPD $\downarrow$ & DAS $\downarrow$ \\
			\midrule
			w/o pixel-level condition     & 0.2343 & 138.45 & 3.32 \\
			w/o semantic condition          & 0.1492 & 141.61 & 3.21 \\
			w/o reconstruction branch            & 0.1588 & 644.48 & 3.84 \\
			w/o cross-modality             & 0.1644 & 98.80 & 3.20 \\
			feature addition condition    & 0.1532 & 220.51 & 3.23 \\
			\midrule
			full model           & \textbf{0.1477} & \textbf{97.49} & \textbf{3.16} \\
			\bottomrule
	\end{tabular}}
	\caption{Ablation studies of MultiEditor modules.}
	\label{tab:ablation}
\end{table}
\subsection{Ablation Studies}
\subsubsection{Multi-Level Appearance Control Mechanism Analysis} 
We evaluate the influence of the multi-level appearance control mechanism on generation performance, with quantitative results summarized in Table~\ref{tab:ablation}. First, removing the pixel-level condition—which prevents explicit priors from 3DGS from being pasted into masked regions—substantially degrades image appearance and point cloud geometry, underscoring the importance of low-level cues for accurate editing. Second, we ablate the semantic condition by excluding global semantic features extracted via CLIP. While the image modality remains relatively unaffected—possibly due to compensation from other fine-grained signals—the quality of point cloud generation noticeably degrades, highlighting the vital role of semantic guidance in textureless 3D synthesis. Lastly, removing the reconstruction branch causes a marked decline in visual and geometric fidelity, confirming its necessity for refining the generated outputs.
\subsubsection{Cross-Modality Module Analysis} 
We ablate the cross-modality condition module to turn off mutual guidance between modalities. As shown in Table~\ref{tab:ablation}, removing this module substantially degrades cross-modality consistency. While the shared 3DGS prior offers some semantic and structural constraints, it alone is inadequate for achieving precise alignment between modalities. Moreover, substituting deformable attention with naive feature addition degrades the model’s capacity to accurately align local structures across modalities, reducing consistency in the generated results.

\subsection{Downstream Task Benefits}
\begin{table}[t]
	\centering
	\small
	\resizebox{1\linewidth}{!}{
		\begin{tabular}{ccccc}
			\toprule
			\multicolumn{5}{c}{\textbf{2D Detection}} \\
			\midrule
			Method & DataType & AP@0.5 $\uparrow$ & AP@0.7 $\uparrow$ & mAP 0.5--0.95 $\uparrow$ \\
			\midrule
			\multirow{2}{*}{Yolo v5} 
			& Real         & 27.0  & 21.2  & 16.9  \\
			& Real + Gen   & 31.5(\textbf{+4.5\%})  & 23.0(\textbf{+1.8\%})  & 18.2(\textbf{+1.3\%})  \\
			\midrule
			\multicolumn{5}{c}{\textbf{3D Detection}} \\
			\midrule
			Method & DataType & AP@0.7 Easy $\uparrow$ & AP@0.7 Mod. $\uparrow$ & AP@0.7 Hard $\uparrow$ \\
			\midrule
			\multirow{2}{*}{PointPillars} 
			& Real         & 43.02 & 34.23 & 28.54 \\
			& Real + Gen   & 48.36(\textbf{+5.34\%}) & 35.45(\textbf{+1.22\%}) & 29.17(\textbf{+0.63\%}) \\
			\bottomrule
		\end{tabular}
	}
	\caption{Detection performance on van-class objects using 2D and 3D detection models trained with real and augmented (real + generated) data.}
	\label{tab:downstream}
\end{table}
To evaluate the effectiveness of MultiEditor in downstream perception tasks, we augment the training samples of the van class in the KITTI dataset to improve detection performance for this underrepresented category. Using YOLOv5 \cite{yolo} and PointPillars \cite{pointpillars} as 2D and 3D detectors, we follow the PointPillars split of the KITTI dataset for training and evaluation. As shown in Table~\ref{tab:downstream}, van-class vehicles are scarce in the training set (with only 1,297 instances), leading to suboptimal detection accuracy. To mitigate this, we select 10 van instances from the 3DRealCar dataset and insert them into training images using MultiEditor. This process yields 1,192 new samples containing van-class vehicles, which are then merged with the original training set. Results show that incorporating MultiEditor-generated samples significantly improves van-class detection accuracy for both 2D and 3D detectors. This demonstrates the practical utility of MultiEditor in alleviating long-tail data imbalance and boosting detection accuracy for rare object classes.
\section{Conclusions}
We present MultiEditor, a novel multimodal object editing framework that leverages 3DGS priors to enable joint editing of images and LiDAR point clouds. By incorporating a dual-branch diffusion architecture, multi-level control mechanisms, and a cross-modality condition module, MultiEditor effectively models the joint distribution across modalities, enabling high-fidelity and consistent scene editing. Extensive experiments demonstrate that our method enables accurate, consistent, and controllable multimodal editing for autonomous driving scenarios.
\bibliography{aaai2026}
\newpage
\section{Overview}
This supplementary material first provides detailed implementation specifics. We then present additional qualitative results to further demonstrate the effectiveness of the proposed method. Lastly, we discuss the limitations of our framework and outline potential directions for future work.

\section{Detailed Implementation Specifics} \label{dis}
In this part, we explain details in the implementation of MultiEditor including the data construction process, training schedule, and evaluation metrics. Our code will be made publicly available upon the acceptance of this paper.
\subsection{Data Construction Process}
As no existing dataset is tailored for joint multimodal object editing, we construct a dedicated dataset based on the widely used KITTI benchmark \cite{kitti}. Specifically, we first apply the Grounding DINO model \cite{groundingdino} for text-guided object detection, followed by the Segment Anything Model (SAM) \cite{sam} to accurately segment target objects and generate corresponding scene masks. Subsequently, we leverage the calibrated geometric transformation between the camera and LiDAR to project the image-space masks onto the LiDAR range image coordinate system. This allows for accurate localization and extraction of the corresponding point clouds and their associated masks.   

Traditional mask-guided image editing methods \cite{driveeditor,mobi} typically restrict the generative scope to within the predefined mask boundaries, limiting their applicability in real-world scenarios where visual elements such as shadows often extend beyond these regions. To address this issue, we enhance the image branch to support unconstrained editing by post-processing the generated image data.

\begin{figure}[t]
	\centering
	\includegraphics[width=0.9\linewidth]{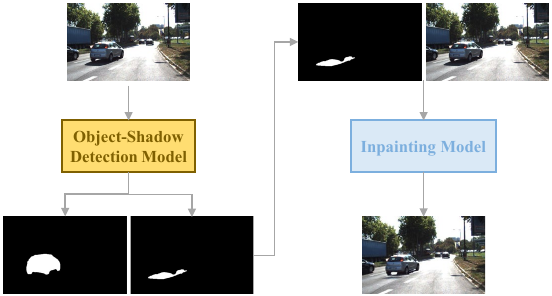}
	\caption{The pipeline of dataset construction. We use the object-shadow detection model \cite{ssis} to predict pairs of object and shadow masks in the real image. Then we apply the inpainting model \cite{ldm} to get a deshadowed image.}
	\vspace{-12pt}
	\label{fig7}
\end{figure}

Specifically, as shown in Figure~\ref{fig7}, we employ the object-shadow detection model \cite{ssis} to extract shadow masks cast by the target objects and utilize a pretrained image inpainting model \cite{ldm} to perform high-quality shadow removal, resulting in clean, shadow-free images suitable for downstream unconstrained editing tasks.

After completing the above inpainting process, we manually screened the generated data to ensure high-quality training samples. Specifically, we removed instances with poor-quality object or shadow masks and those exhibiting visible artifacts in the inpainted regions. This rigorous filtering yields a high-quality dataset of 4,021 image–point cloud pairs for training and 1,256 for testing.
\begin{figure*}[t]
	\centering
	\includegraphics[width=0.9\linewidth]{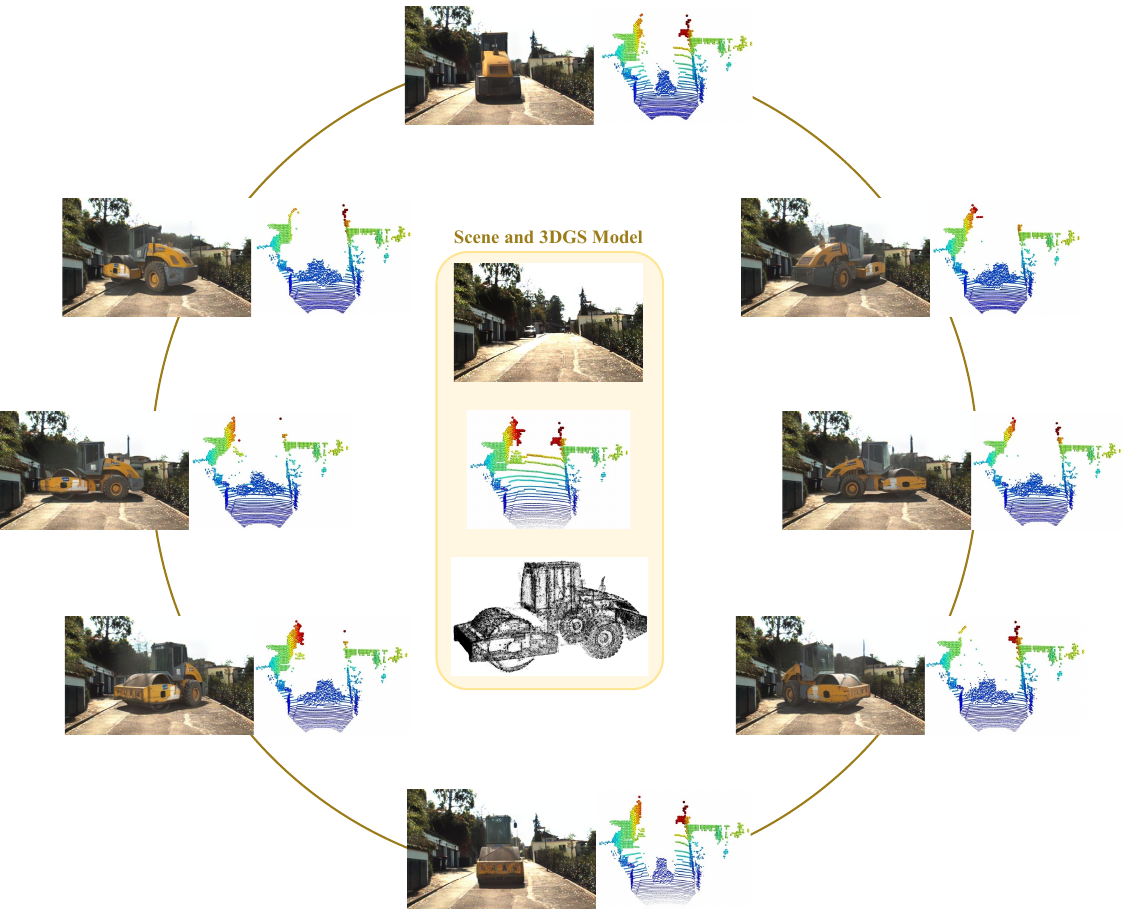}
	\caption{MultiEditor demonstrates high controllability and flexible editing of complex-shaped vehicles. A roller vehicle is inserted into the scene at $45^\circ$ intervals, showcasing consistent and precise object editing performance.}
	\label{fig8}
\end{figure*}
\begin{figure*}[t]
	\centering
	\includegraphics[width=0.9\linewidth]{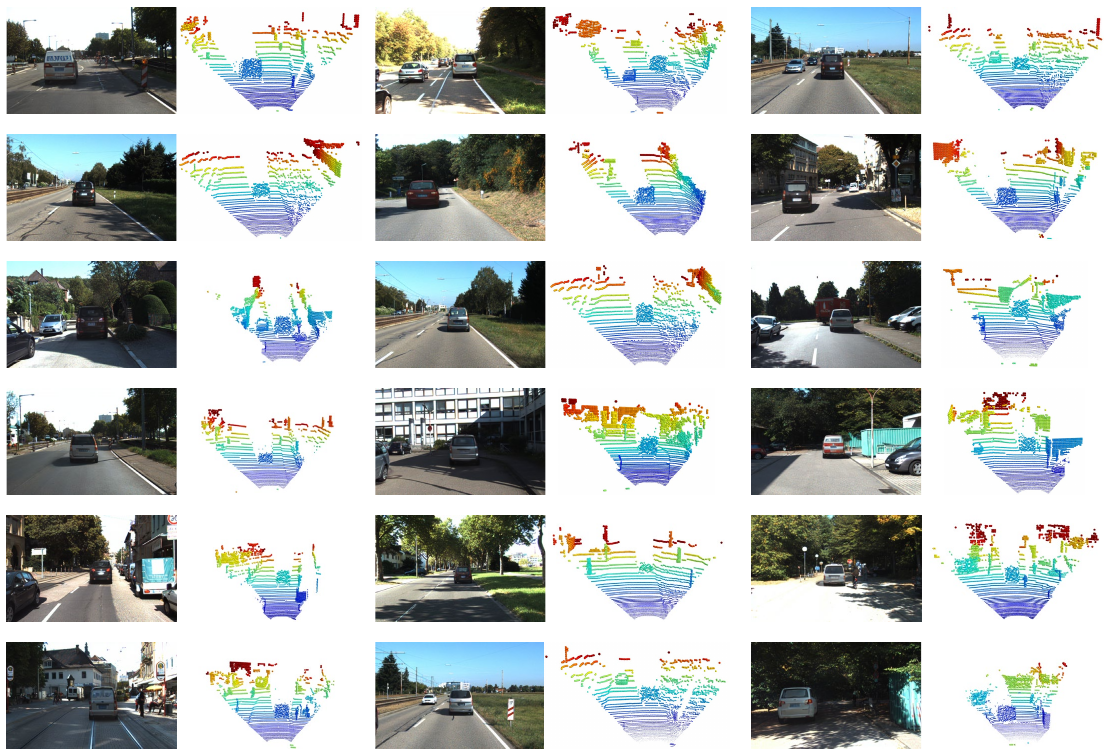}
	\caption{Multimodal data generation for downstream tasks. We insert van-class vehicles at varying poses and distances into image and point cloud modalities. This enhances the diversity of training data and improves the performance of 2D and 3D detectors in recognizing van-class objects.}
	\vspace{-12pt}
	\label{fig10}
\end{figure*}
\subsection{Training Schedule}
The overall training pipeline of MultiEditor comprises five stages:
\begin{itemize}
	\item \textbf{Stage 1:} We train a variational autoencoder (VAE) from scratch for the range image modality in the point cloud branch.
	
	\item \textbf{Stage 2:} The range image VAE is frozen, and a latent diffusion model is trained on its latent representations.
	
	\item \textbf{Stage 3:} The image VAE is frozen, and a latent diffusion model is trained on unprocessed (i.e., shadow-containing) images. This enables object insertion within masked regions, albeit still constrained to the mask.
	
	\item \textbf{Stage 4:}  We fine-tune the diffusion model from Stage 3 using only the refinement strategy in the image branch, with training performed on deshadowed image data. This encourages the model to learn generation distributions beyond the mask boundaries, enabling unconstrained editing capability.
	
	\item \textbf{Stage 5:} Finally, we perform end-to-end joint training of the entire multimodal framework based on the module trainable states defined in Figure 2. We aim to improve cross-modal consistency and overall editing performance further.
\end{itemize}
During all training stages, we use four NVIDIA L20 GPUs. In Stage 1, the VAE for LiDAR range images is trained with a batch size of 2 and a learning rate of 4.5e-5 for 40 epochs, with the discriminator activated after 1000 iterations. In Stage 2, the LiDAR LDM is trained with a batch size of 2 and a learning rate 4.0e-5 for 100 epochs. In Stage 3, the image LDM is trained with a batch size of 2 and a learning rate 1.0e-5 for 40 epochs. In Stage 4, we fine-tune the image LDM from Stage 3 using deshadowed images, with a batch size of 2 and a learning rate of 1.0e-5 for 160 epochs. Finally, Stage 5 performs end-to-end training of the whole multimodal framework with a batch size of 1 and a learning rate 2.0e-5 for 60 epochs. For ablation studies, we adopt the same complete training configuration as described above. 

In terms of image size configuration, following the experimental setup of SGD \cite{sgd}, we apply a center crop of size $600\times375$ to the input images. Correspondingly, the range images of point clouds are cropped based on the field of view (FOV) of the cropped images and resized to $128\times64$. For downstream tasks, the cropped image and point cloud patches are seamlessly stitched back into the original data.

To improve robustness, we perform data augmentation on the target objects. For the image branch, we apply RandomBrightnessContrast, Rotate, HueSaturationValue, Blur, and ElasticTransform, each with a 20\% probability. For the point cloud branch, only the Rotate augmentation is used, also with a 20\% probability.

In addition, the value of $\lambda_{refine-C}$ in Equation (11) of the main text is set to 0.01.
\subsection{Evaluation Metrics}
In our quantitative evaluation, we apply the FID metric and LPIPS metric from DriveEditor \cite{driveeditor}, CLIP-I
metric from AnyDoor \cite{anydoor}, CD metric from PVD \cite{cd}, and FPD metric from R2DM \cite{r2dm}. For the DAS metric, we follow the protocol of X-DRIVE \cite{xdrive} by applying a pretrained DepthAnythingV2 model \cite{depth} with a ViT-B backbone \cite{dosovitskiy2020image} to the synthesized images for depth estimation. The estimated depth is then rescaled to an absolute scale using ground-truth LiDAR point clouds. Finally, we project the generated point clouds into the image space to obtain sparse depth values and compute the mean absolute error (MAE) against the estimated depth map as our DAS score.

\section{More Visualization Results}
\subsection{More Editing Results}
As shown in Figure~\ref{fig8}, we demonstrate the high-fidelity and flexible editing capabilities of MultiEditor on atypical vehicles. Specifically, we reconstruct a 3D Gaussian Splatting (3DGS) representation of a roller vehicle from the 3DRealCar dataset \cite{3drealcar}. RGB images and depth maps are rendered at $45^\circ$ intervals, and the target object is inserted into the scene using MultiEditor. Figure~\ref{fig8} shows generation results across different viewpoints, highlighting MultiEditor’s strong controllability and geometric consistency in editing complex driving scenes.
\subsection{Visualization of Generated Data for Downstream Tasks}
To evaluate the effectiveness of our generated data in downstream perception tasks, we augment existing driving scenes by inserting van-class vehicles with diverse poses and distances into both the image and point cloud modalities, as shown in Figure~\ref{fig10}. This augmentation addresses the long-tail distribution problem commonly found in detection datasets, where van-class objects are underrepresented. By enriching the training set with high-quality, multimodal synthetic samples, we observe improved performance in 2D and 3D object detectors. The inserted objects are geometrically and photometrically consistent with the surrounding environment, demonstrating the practical utility of MultiEditor in real-world data enhancement scenarios.
\subsection{The Impact of the Shadow Refinement Training Stage}
\begin{figure}[t]
	\centering
	\begin{subfigure}{0.48\linewidth}
		\centering
		\includegraphics[width=\linewidth]{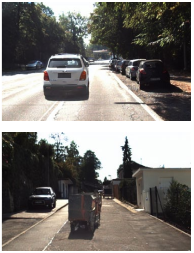}
		\caption{Without the shadow refinement stage.}
		\label{fig3a}
	\end{subfigure}
	\begin{subfigure}{0.48\linewidth}
		\centering
		\includegraphics[width=\linewidth]{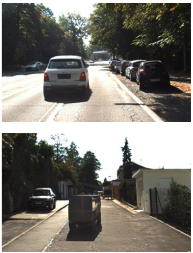}
		\caption{With the shadow refinement stage.}
		\label{fig5b}
	\end{subfigure}
	\caption{Visualization results with and without the shadow refinement stage.}
	\label{fig9}
	\vspace{-12pt}
\end{figure}
As shown in Figure~\ref{fig9}, we evaluate the effect of the shadow refinement stage on editing quality. The left column shows results without refinement, where inserted objects suffer from lighting inconsistency, unnatural shadows, and poor scene integration. In contrast, the right column presents results with shadow refinement, leading to more coherent illumination, improved shadow consistency, and more realistic blending with the environment.
\section{Limitation and Future Works}
This work focuses on the joint editing of consistent multimodal data. Due to limited computational resources, we leave several natural extensions of the MultiEditor framework for future exploration. First, our current implementation supports editing only a single-frame image and the corresponding LiDAR point cloud within the same field of view (FOV). A straightforward extension would support multi-view image inputs combined with a single-frame LiDAR scan. Second, integrating temporal attention mechanisms, such as those proposed in Panacea \cite{panacea}, into MultiEditor could enable long-sequence multimodal, controllable, and temporally consistent video editing.
\end{document}